# Predictive Maintenance for Industrial IoT of Vehicle Fleets using Hierarchical Modified Fuzzy Support Vector Machine

Arindam Chaudhuri
Samsung R & D Institute Delhi
Noida – 201304 India
arindamphdthesis@gmail.com

**Abstract:** Connected vehicle fleets are deployed worldwide in several industrial IoT scenarios. With the gradual increase of machines being controlled and managed through networked smart devices, the predictive maintenance potential grows rapidly. Predictive maintenance has the potential of optimizing uptime as well as performance such that time and labor associated with inspections and preventive maintenance are reduced. In order to understand the trends of vehicle faults with respect to important vehicle attributes viz mileage, age, vehicle type etc this problem is addressed through hierarchical modified fuzzy support vector machine (HMFSVM). The proposed method is compared with other commonly used approaches like logistic regression, random forests and support vector machines. This helps better implementation of telematics data to ensure preventative management as part of the desired solution. The superiority of the proposed method is highlighted through several experimental results.

## 1. Introduction

The present business setup has connected vehicles [1] which forms an integral part of operations in every industry today. Whether in a factory, fleet, hospital, office or home these vehicles are producing huge amount of data which is projected to grow steadily in the coming years. Today organizations are struggling to capture and harness the power of IoT information [2] so they can apply operational insights regarding the devices and get ahead of unplanned downtime. Organizations are always looking for faster ways [3] to realize the sensor information and transform it into predictive maintenance insights [4]. Only 10% of today's IoT-generated data is used for deeper analysis. The common approach is to amass the sensor data and treat this as Big Data problem. But this misses Moore's Law alteration of the landscape of devices.

The cost of vehicle downtime is significant towards customers' demand for higher up-times and aggressive service level agreements (SLAs). The service providers look for predictive maintenance techniques which use accurate real-time vehicle information. This helps them to determine the vehicle's condition as and when maintenance is to be done. This approach provides savings [5] in terms of cost over traditional preventive maintenance tasks. The actual value of predictive maintenance allows corrective maintenance scheduling and unexpected vehicle failures prevention. The success lies in getting the right information at the right time. The maintenance work can be better planned when it is known which vehicle needs attention beforehand. With connected vehicles predictive maintenance solution [6] the users achieve timely maintenance towards increased up-times, better plan maintenance in reducing unnecessary field service calls, optimizing repair parts replacement, reducing unplanned stops, improving vehicle performance and service compliance reporting. Nowadays customers connect their products for analysing vehicle data enabling preventative maintenance. These customers implement business rules and integrate alarms into enterprise business systems towards automating field service, repair parts deployment as well as other preventative maintenance tasks.

Vehicles are designed with temperature, infrared, acoustic, vibration, battery-level and sound sensors in order to monitor conditions which can form as initial maintenance indicators as shown in Figure 1. The service needs are determined through inputs from these sensors. The predictive maintenance programs are driven by customers using cloud which helps them to collect and manage vehicle data alongwith visualization and analytics [7] tools to make better decisions. Most of the IoT analytics tools have characteristics like real-time IoT big data visual exploration, instant visual event trending snapshots and key performance indicators insights, visualization of live and historical data apps mashed with IoT data and dashboards which users can assemble and share.

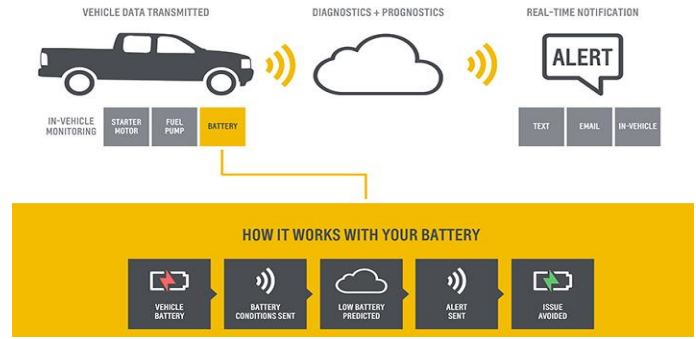

**Figure 1: Predictive maintenance of vehicles**

To solve the problems posed by predictive maintenance of vehicle fleets in a specified garage for telecom-based company statistical and machine learning techniques [8] are used to develop real time analytics solutions. In this direction hierarchical modified fuzzy support vector machine (HMFSVM) is proposed here to achieve the predictive analytics task. The predictive analytics problem comprises of the classification problem which is performed through HMFSVM, a variant of MFSVM and FSVM developed by Chaudhuri et al [9], [10], [11]. In MFSVM success of classification is in appropriate selection of the fuzzy membership function [12]. The fuzzy membership function is articulated in terms of center and radius of each class in feature space with kernel. When training samples are nonlinear, the input space is represented as high dimensional feature space. This helps in computing separating surface through linear separating method [13], [14]. The membership function represents the importance towards sampling the decision surface. The larger the fuzzy membership function's value, the importance of point increases. For this classification problem, the vehicle data is paralysed with population drift where the customer behavior changes over time. To support this hypothesis, an appropriate model is developed based on data adapted from narrow time period where the variability likelihood is less. Based on the success of MFSVM in financial applications [9], it is hierarchically extended to HMFSVM. HMFSVM gives better sensitivity to class imbalance problem, number of support vectors reduction, decreases training times, model complexity with better generalization. The experimental results support HMFSVM's superiority over algorithms.

Some of the key tasks performed in this analysis include [11] processing the data for prediction, data cleaning, data analysis, model building, model training and model testing and validation. The HMFSVM based predictive analytics engine correlates parameters like vehicle make description, vehicle model, vehicle age, distance travelled, past maintenance history etc using regression analysis and predicts probability that a certain vehicle will require maintenance or repair within a certain time period. These solutions provide necessary infrastructure to rapidly deliver high value predictive insights that reduces waiting times of the vehicles due to different failures. The benefits to the vehicle fleets includes breakdowns avoidance, reduction in the number of garage visits, reduced cost, increased period of leasing and increased availability of the vehicles. The value of implementing an IoT strategy for predictive maintenance using HMFSVM for advanced analytics is that maintenance becomes real-time, dependable, secure and truly predictive. By using real time analytics to perform timely maintenance on vehicles organizations can move away from reactive maintenance. Predictive maintenance maximizes the lifespan of connected vehicles and ensures optimized production using actionable, advanced visual analytics to keep watchful, predictive eye on operations. In heavy industries unanticipated vehicle component breakdown could have catastrophic effects on the overall system. Additionally, the resulting unplanned downtime could be hours, days or even weeks which depends on the problem's severity and the complexity in returning towards normal operations. This research highlights the tremendous value in harnessing the vehicle data as part of an overall IoT strategy such that the problems can be seen well before they get to the point of severity. Visual insight and predictive analytics not only help avoid unanticipated failures but also optimize overall operations while significantly reducing overall resource and repair costs.

This article is structured as follows. The section 2 presents the datasets used for performing the experiments. In section 3 some of the traditional approaches to vehicle maintenance are presented. The section 4 provides some insights on data preparation. The prediction hypothesis is highlighted in section 5. The sections 6 and 7 highlight

MFSVM and HMFSVM respectively. In section 8 input output variables mapping visa vis HMFSVM is presented. The experimental results and challenges faced during predictive model hypothesis are highlighted in section 9. Finally in section 10 conclusions are given.

## 2. Available Datasets for Prediction

The datasets are adopted from telecom-based company [8] from a specified garage. The datasets available for predictive analytics include data from 5 garages. The data from the first garage spanned over a period of 14 months and 24 months for other garages. The number of data records available from first garage was 3923 of which 2097 was used for prediction analysis. The number of data records available from other garages was 890665 of which 11456 was used for prediction analysis. The reasons behind not using the entire available data for prediction include: (a) there were data rows with invalid values (null, value out of range, blanks etc) which have been considered as outliers and have been removed (b) vehicle repair types count (audits, automatic transmission, body exteriors, body interiors etc) did not have considerable values for all the vehicles were removed (c) fuel economy score and driver behavior score were available only for the first garage which was not used because it reduces the prediction accuracy. There was great degree of skewedness in the training data. Before applying the data to the prediction engine, it was balanced through external imbalance learning method. The dataset was used after the majority class under-sampling which makes the number of good training samples equal to the rejected training examples.

## 3. Traditional Approaches to Vehicle Maintenance

Companies have sent field technicians towards regular diagnostic inspections and preventive maintenance [8] as per prespecified schedule of vehicles. Here an appreciable amount of cost and labor is involved with little assurance that failure would not happen in inspections. Despite the traditional preventive measures put in place to avoid vehicle downtime, asset maintenance is often rife with unexpected failure which requires unplanned maintenance. Traditional checklists or quality control procedures provide insight but only after the vehicles have begun to show signs of failure or have already failed.

Preventive maintenance is required by most organizations [15] but with schedules that have been predetermined by vehicle manufacturers. These schedules often lead to replacement of repair parts well before they are required. Unnecessary part replacement and unrequired servicing creates tremendous expenses [16] especially considering the fact that maintenance may not be based on any operational data preventive analysis.

When unforeseen vehicle failures occur organizations typically move to a reactive maintenance model. This is prohibitively expensive and substantially inefficient. Due to lack of information service technicians are left scrambling to limit operational losses and bring vehicle back online. Predictive maintenance using advanced analytics catches the issues early and at the source to identify and address potential vehicle issues and failures before they happen. It also optimizes the end-to-end service operations and results in an overall improvement in vehicle reliability and performance.

## 4. Data Preparation and Mining

The experimental data was prepared [11] by recognizing the base and derived variables from the available data. The base variables were directly available in the source data [8] and includes vehicle registration date, vehicle registration number, odometer reading etc. The derived variables were created as calculated fields from the base variables fields in order to provide additional meaning to the causes of repairs and faults to allow the prediction algorithm arrive at a robust prediction. Some of derived variables created for this analysis includes age of the vehicle, count of occurrence of each service type, average labor time and average cost of spare parts. The base and derived variables that were inputs to the model to arrive at predictions constituted the predictor variables. The data cleaning was performed on the predictor variables [9] where outliers viz extreme values, null and blank values, incorrect values etc were removed.

## 5. Predictive Maintenance Prediction Hypothesis

In this section the predictive maintenance predictive hypothesis is highlighted. The predictive analysis was based on the following prediction hypothesis [8]:

1. The higher number of visits to the garage indicates higher likelihood of repeat visits
2. The higher the age of the vehicle indicates greater probability of needing maintenance
3. The higher odometer reading in vehicle indicates greater wear and tear probability and hence need for maintenance
4. The greater number of occurrences of repair type (work area description) for specific vehicle indicates higher probability of the recurrence
5. The higher average labour cost over 1 year suggests vehicle involves high cost of maintenance and will need frequent servicing
6. If the number of tasks performed in the last job is high the vehicle potentially has many issues and is more likely to have faults
7. The higher repair parts cost indicate there is a higher wear and tear on the vehicle and hence likelihood of other parts failing
8. If there is high labor time in the last Job there could be larger problem in the vehicle and hence probability of vehicle needing further service is higher

## 6. Modified Fuzzy Support Vector Machine

Both FSVM [10] and MFSVM [9] have been used in bankruptcy prediction and credit approval classification respectively. A brief overview of FSVM and MFSVM [9] proposed by Chaudhuri et al is presented here. Incorporating fuzzy membership function to SVM [12] gives birth to FSVM. In classical SVM, each sample point is fully assigned to either of the two classes. In many applications there are points which may not be exactly assigned to either of the two classes. Here each point does not bear similar significance to the decision surface.

### 6.1 Mathematical Foundations

To address this problem, fuzzy membership is assigned to each input point of SVM such that the input points contribute uniquely towards decision surface construction. The corresponding input's membership is reduced such that total error term contribution is decreased. Each input point is treated with higher membership as an input of opposite class. The new fuzzy machine to makes full data use and achieves better generalization ability. Let us suppose training sample points are considered as:

$$Sample\_Points = \{(S_i, y_i, sp_i); i = 1, \ldots\ldots, P\} \quad (1)$$

Here each $S_i \in R^N$ is training point and $y_i \in \{-1, +1\}$ represent label of the class; $sp_i; i = 1, \ldots\ldots, P$ represents fuzzy membership function such that $p_j \leq sp_i \leq p_i; i = 1, \ldots\ldots, P$ where $p_j > 0$ and $p_i < 1$ are sufficiently small constants. We denote set $S = \{S_i | (S_i, y_i, sp_i) \in Sample\_Points\}$ containing two classes. One class contains point $S_i$ with $y_i = 1$ represented through $Class^+$:

$$Class^+ = \{S_i | S_i \in Sample\_Points \land y_i = 1\} \quad (2)$$

Other class contains point $S_i$ with $y_i = -1$ represented through $Class^-$:

$$Class^- = \{S_i | S_i \in Sample\_Points \land y_i = -1\} \quad (3)$$

Obviously $Point\_Space = Class^+ \cup Class^-$. The quadratic classification proposition is represented as:

$$\min \frac{1}{2} \|\omega\|^2 + C \sum_{i=1}^{P} sp_i \rho_i \quad (4)$$

subject to
$$y_i(\omega^T \Phi(S_i) + a) \geq 1 - \rho_i, i = 1, \ldots, P$$
$$\rho_i \geq 0$$

In equation (4) $C$ is regularization parameter. The fuzzy membership $sp_i$ governs behaviour of sample point $S_i$ towards one class and $\rho_i$ measures the error in SVM. The factor $sp_i \rho_i$ is error measure considering different weights. The smaller $sp_i$ reduces parameter $\rho_i$ effect in equation (4) such that point $S_i$ is less significant. The above quadratic proposition can also be solved through corresponding dual problem [17]. Given the training points sequence in equation (4), the mean of class $Class^+$ and $Class^-$ is $mean_+$ and $mean_-$ respectively. The class $Class^+$ radius is:

$$radius_+ = \max\|mean_+ - S_i\|; S_i \in Class^+ \quad (5)$$

The radius of class $Class^-$ is:

$$radius_- = \max\|mean_- - S_i\|; S_i \in Class^- \quad (6)$$

The fuzzy membership $sp_i$ [18] is given by:

$$sp_i = \begin{cases} 1 - \frac{\|mean_+ - S_i\|^2}{(radius_+ + \theta)^2} & if\ S_i \in Class^+ \\ 1 - \frac{\|mean_- - S_i\|^2}{(radius_- + \theta)^2} & if\ S_i \in Class^- \end{cases} \quad (7)$$

With the above membership function FSVM achieves good performance. A particular sample in training set contributes minimum to final classification result and outliers' effect are eliminated through samples' average.

Now MFSVM is formulated based on FSVM. Consider sample $S_i \in Sample\_Space$. Let $\Phi(S_i)$ represent mapping function through input into feature space. The hyperbolic tangent kernel is $Kernel(S_i, S_j) = \tanh[\Phi(S_i).\Phi(S_j)]$ [19], [20]. Hyperbolic tangent kernel represented in Figure 2 is also known as sigmoid kernel. In [21] kernel matrix of hyperbolic tangent function is not positive semi definite for certain values. More insights are given by Burges et al [13], [22]. The sigmoid kernel has been used with appreciable success [23]. This fact motivates us in using this kernel in this work.

Now $\Phi_+$ is defined as the center of class $Class^+$ as:

$$\Phi_+ = \frac{1}{samples_+} \sum_{S_i \in Class^+} \Phi(S_i) freq_i \quad (8)$$

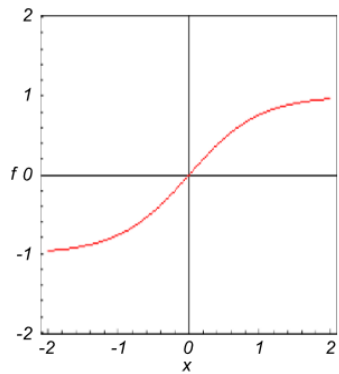

**Figure 2: Graphic representation of hyperbolic tangent kernel for real values**

In equation (8) $samples_+$ is the number of samples of class $Class^+$ with frequency $freq_i$ of $i^{th}$ sample in $\Phi(S_i)$. Again $\Phi_-$ is the class center of $Class^-$ and is defined as:

$$\Phi_- = \frac{1}{samples_-}\sum_{S_i \in Class^-} \Phi(S_i)\, freq_i \qquad (9)$$

In equation (9) $samples_-$ is the number of samples of class $Class^-$ with frequency $freq_i$ of $i^{th}$ sample in $\Phi(S_i)$. The $Class^+$ radius is given by:

$$radius_+ = \frac{1}{n}\max\|\Phi_+ - \Phi(S_i)\| \qquad (10)$$

Here $S_i \in Class^+$ and the radius of $Class^-$ is given by:

$$radius_- = \frac{1}{n}\max\|\Phi_- - \Phi(S_i)\| \qquad (11)$$

Here $S_i \in Class^-$ and $n = \sum_i freq_i$ such that we have:

$$radius_+^2 = \frac{1}{n}\max\|\Phi(S') - \Phi_+\|^2$$
$$radius_+^2 = \frac{1}{n}\max[\Phi^2(S') - 2\Phi(S')\Phi_+ + \Phi_+^2]$$
$$radius_+^2 = \frac{1}{n}\max\left[\Phi^2(S') - \frac{2}{samples_+}\sum_{S_i \in Class^+}\tanh[\Phi(S_i).\Phi(S')] + \frac{1}{samples_+^2}\sum_{S_i \in Class^+}\sum_{S_j \in Class^+}\tanh[\Phi(S_i).\Phi(S_j)]\right]$$
$$radius_+^2 = \frac{1}{n}\max\left[Kernel(S',S') - \frac{2}{samples_+}\sum_{S_i \in Class^+} Kernel(S_i,S') + \frac{1}{samples_+^2}\sum_{S_i \in Class^+}\sum_{S_j \in Class^+} Kernel(S_i,S_j)\right] \qquad (12)$$

In equation (12) $S' \in Class^+$ and $samples_+$ is number of training samples in $Class^+$.

$$radius_-^2 = \frac{1}{n}\max\left[Kernel(S',S') - \frac{2}{samples_-}\sum_{S_i \in Class^-} Kernel(S_i,S') + \frac{1}{samples_-^2}\sum_{S_i \in Class^-}\sum_{S_j \in Class^-} Kernel(S_i,S_j)\right] \qquad (13)$$

In equation (13) $S' \in Class^-$ and $samples_-$ is number of training samples in $Class^-$. The distance square between sample $S_i \in Class^+$ and its class center is given as:

$$distance_{i+}^2 = \|\Phi(S_i) - \Phi_+\|^2 = \Phi^2(S_i) - 2\tanh[\Phi(S_i)\Phi_+] + \Phi_+^2$$
$$distance_{i+}^2 = Kernel(S_i,S_j) - \frac{2}{samples_+}\sum_{S_j \in Class^+} Kernel(S_i,S_j) + \frac{1}{samples_+^2}\sum_{S_j \in Class^+}\sum_{S_k \in Class^+} Kernel(S_j,S_k) \qquad (14)$$

Similarly, the distance square between sample $S_i \in Class^-$ and its class center is given as:

$$distance_{i-}^2 = Kernel(S_i,S_j) - \frac{2}{samples_-}\sum_{S_j \in Class^-} Kernel(S_i,S_j) + \frac{1}{samples_-^2}\sum_{S_j \in Class^-}\sum_{S_k \in Class^-} Kernel(S_j,S_k) \qquad (15)$$

Now $\forall i;\ i = 1,\ldots,P$ the fuzzy membership function $sp_i$ is:

$$sp_i = \begin{cases} 1 - \sqrt{\dfrac{\left\|distance_{i+}^2\right\|^2 - \left\|distance_{i+}^2\right\|\,radius_+^2 + radius_+^2}{\left(\left\|distance_{i+}^2\right\|^2 - \left\|distance_{i+}^2\right\|\,radius_+^2 + radius_+^2\right) + \epsilon}} & if\ y_i = 1 \\[2ex] 1 - \sqrt{\dfrac{\left\|distance_{i-}^2\right\|^2 - \left\|distance_{i-}^2\right\|\,radius_-^2 + radius_-^2}{\left(\left\|distance_{i-}^2\right\|^2 - \left\|distance_{i-}^2\right\|\,radius_-^2 + radius_-^2\right) + \epsilon}} & if\ y_i = -1 \end{cases} \qquad (16)$$

In equation (16) $\epsilon > 0$ such that $sp_i$ is not zero and is function of center and radius of each class in feature space represented through kernel. The training samples are either linear or nonlinear separable. This method accurately represents each sample point contribution towards separating hyperplane construction in feature space [24]. This helps MFSVM to reduce outliers' effect more efficiently with better generalization ability.

# 7. Hierarchical Modified Fuzzy Support Vector Machine for Predictive Maintenance

Now hierarchical version of MFSVM is presented. The prediction model was build using HMFSVM. This is a decision-based model that produces predictions as probability values. The computational benefits accrued from MFSVM serve the major prima face. HMFSVM supersedes MFSVM considering similarity-based classification accuracy and running time as data volume increases [25]. The proposed model's architecture is shown in the Figure 3 with temporal data sequences modeled through MFSVMs. These are combined together towards HMFSVM. HMSVM architecture represents data behavior across multiple relevant features.

The model is composed of 6 layers. At 1st and 2nd layers relatively small MFSVMs are utilized. The number of MFSVMs are increased at layers 3, 4 and 5. MFSVMs in last layer are constructed over subset of examples in 5th layer is Best Matching Unit (BMU). MFSVMs at last layer can be larger than used in 1st to 5th layers. It improves resolution and discriminatory capacity of MFSVM with less training overhead. Building HMFSVM requires several data normalization operations. This provides for initial temporal pre-processing and inter-layer quantization between 1st to 5th layers. The pre-processing provides suitable representation for data and supports time-based representation. The 1st layer treats each feature independently with each data instance mapped to sequential values. In case of temporal representation standard MFSVM has no capacity to recall histories of patterns directly. A shift-register of length $l$ is employed in which tap is taken at predetermined repeating interval $k$ such that $l \% k = 0$ where % is modulus operator. The 1st level MFSVMs only receive values from shift register. Thus as each new connection is encountered (at left), content of each shift register location is transferred one location (to right) with previous item in $l^{th}$ location being lost. In case of $n$-feature architecture it is necessary to quantize number of neurons between 1st to 5th level MFSVMs. The purpose of 2nd to 5th level MFSVM is to provide an integrated view of input feature specific MFSVMs developed in 1st layer. There is potential in 2nd to 5th layer MFSVM to have an input dimension defined by total neuron count across all 1st layer MFSVMs. This is brute force solution that does not scale computationally. Given the ordering provided by MFSVM neighboring neurons respond to similar stimuli. The structure of each 1st layer SVM is quantized in terms of fixed number of units using potential function classification algorithm [8]. This reduces number of inputs in 2nd to 5th layers of MFSVM. The units in 4th layer acts as BMU for examples with same class label thus maximizing detection rate and minimizing false positives. However there is no guarantee for this. In order to resolve this 4th layer SVM that act as BMU for examples from more than one class are used to partition data. The 5th layer MFSVMs are trained on subsets of original training data. This enables size of 5th layer MFSVMs to increase which improves class specificity while presenting reasonable computational cost. Once training is complete 4th layer BMUs acts to identify which examples are forwarded to corresponding 5th layer MFSVMs on test dataset. A decision rule is required to determine under what conditions classification performance of BMU at 4th layer RFSOM is judged sufficiently poor for association with 5th layer MFSVM. There are several aspects that require attention such as minimum acceptable misclassification rate of 4th layer BMU relative to number of examples labeled at 4th layer BMU and number of examples 4th layer BMU represent. The basic implication is that there must be optimal number of connections associated with 4th layer BMU for training of corresponding 5th layer MFSVM and misclassification rate over examples associated with 4th layer BMU exceeds threshold.

HMFSVM is represented as success probability towards recovering true hierarchy $Cs^*$ with runtime complexity. Some constraints are subjected on similarity function $S$ with respect to similarities that agree hierarchically considering random noise:

S1 For each $y_i \in Cs_j \in Cs^*$ and $j \prime \neq j$:

$$\min_{y_p \in Cs_j} \mathbb{Exp}[S(y_i, y_p)] - \max_{y_p \in Cs_{j\prime}} \mathbb{Exp}[S(y_i, y_p)] \geq \gamma > 0$$

Here expectations are calculated considering the possible noise on $S$.

S2 For each $y_i \in Ct_j$, a set of $V_j$ words of size $v_j$ drawn uniformly from $Cs_j$ satisfies:

$$\text{Prob}\left(\min_{y_p \in Cs_j} \mathbb{E}\text{xp}[S(y_i, y_p)] - \sum_{y_p \in V_j} \frac{S(y_i, y_p)}{v_j} > \epsilon\right) \leq 2e^{\left\{\frac{-2v_j\epsilon^2}{\sigma^2}\right\}}$$

Here $\sigma^2 \geq 0$ parameterizes noise on $S$. Set $V_{j'}$ of size $v_{j'}$ drawn uniformly from class $Cs_{j'}$ with $j \neq j'$ satisfies:

$$\text{Prob}\left(\sum_{y_p \in V_{j'}} \frac{S(y_i, y_p)}{v_{j'}} - \max_{y_p \in C_{j'}} \mathbb{E}\text{xp}[S(y_i, y_p)] > \epsilon\right) \leq 2e^{\left\{\frac{-2v_{j'}\epsilon^2}{\sigma^2}\right\}}$$

The condition $S1$ states that similarity from character $y_i$ to its class in expectation is larger than similarity from that character in other class. This relates to tight classification condition **[25]**. The condition $S2$ states that within- and-between-class similarities shift away from the other. Considering feature learning stacked MFSVMs extracts data sequences temporal features.

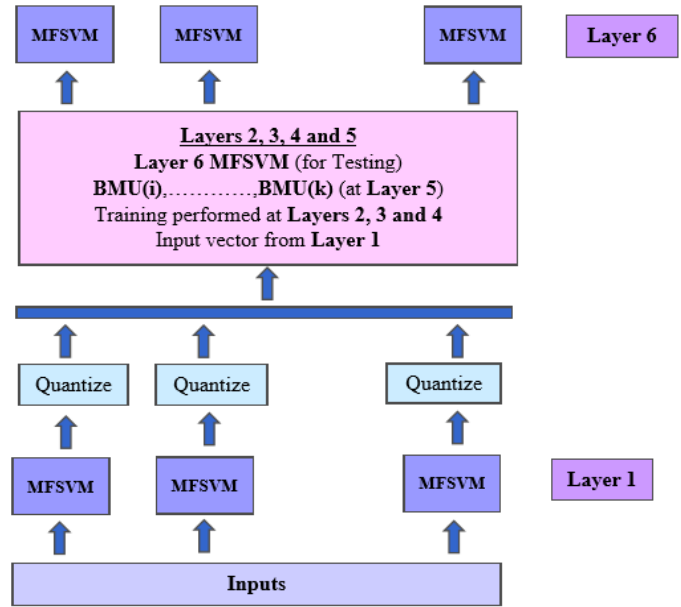

**Figure 3: The architecture of proposed HMFSVM**

## 8. Input Output variables Mapping visa vis to HMFSVM

The input vector to HMFSVM constituted of the base and derived input variables. The base input variables included the vehicle registration number, registration date, number of garage visits of the vehicle, vehicle repair history, odometer reading, repair type of the vehicle and vehicle repair types count. The derived input variables included age of the vehicle, count of occurrence of each service type, average labor time and average cost of spare parts. The input vector was fed to MFSVM at layer 1 and the results were quantized. The quantized results were passed onto layers 2, 3 and 4 where the training was performed repeatedly. After successful training the BMU was found at layer 5. Finally in layer 6 the testing is performed where the predicted probability values are achieved which represents the chance of the vehicle's visit to the garage.

## 9. Experimental Results and Analysis

Here HMFSVM efficiency is highlighted through experimental results on telecom-based company data [8]. A comparative analysis of HMFSVM is also performed with MFSVM, FSVM and other methods like SVM, logistic regression and random forest algorithms. As mentioned is earlier section predictions are produced as probability values by these models. These algorithms run simultaneously on the experimental data and the output of the algorithm that gives the best prediction accuracy is chosen. HMFSVM is also trained on other available garages data alongwith the first garage data. This constitutes the learning phase. The computational workflow constitutes the processed data comprising of the base and derived variables which is fed into HMFSVM. During training the algorithm learns the various inherent patterns in the data. The training process is done iteratively on the training data to fine tune the learning process. HMFSVM testing is performed on the first garage data to create the prediction results. The results are achieved as probability percentage which translates to a time in the future when vehicles have to come up for servicing. The probabilities are classified as follows and highlighted in the Table 1 as (a) Immediate Risk: The probability above 60% (vehicle is in bad shape and has to be sent to garage immediately) (b)Short Term Risk: The probability between 40% and 60% (vehicle could have problems in near future) and (c) Longer Term Risk: The probability less than 40% (vehicle is in a fair condition).

| Classification Category | Number of Vehicles |
|---|---|
| Immediate Risk | 30 |
| Short Term Risk | 175 |
| Longer Term Risk | 1668 |
| **Total Number of Vehicles** | 1873 |

Table 1: Classification results of vehicles

The prediction results obtained are validated with respect to the available results. Keeping in view the classification results sample vehicles from the immediate risk category are considered. Considering an immediate risk vehicle number KE55KZB, the total number of jobs is 13, the total labor hours is 121.35 and the average labor hours is 9.34. These results are shown in Figure 4. Hence it is required to allocate approximately 9.34 hours for this vehicle when scheduling appointment. The typical services required for this vehicle are shown in Figure 5. As obvious from the figure the typical services required for this vehicle are vehicle movement, routine service and electrical system repairs.

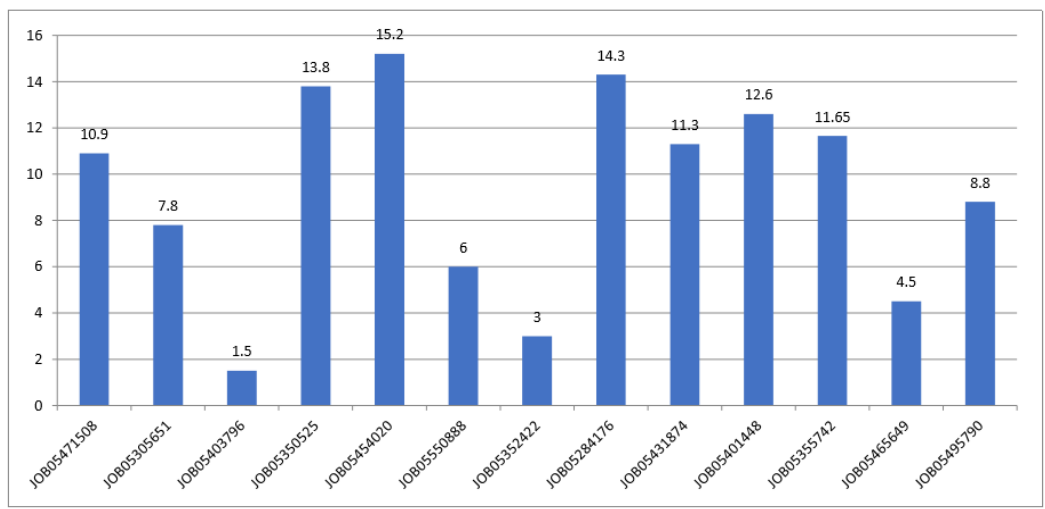

Figure 4: Immediate risk vehicle number KE55KZB (total number of jobs and labor hours)

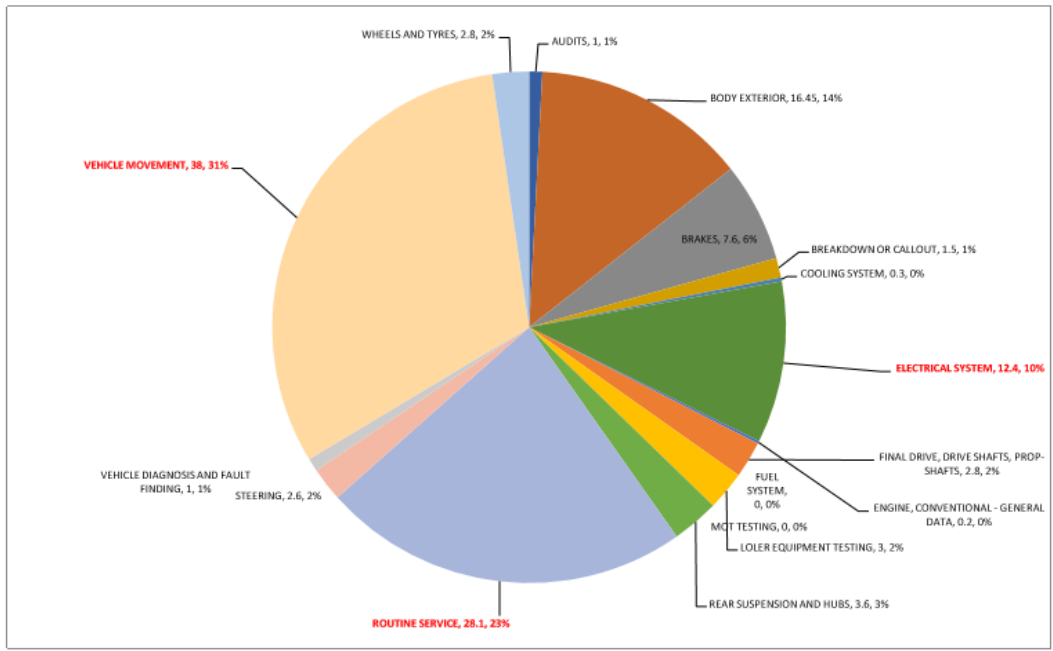

**Figure 5: The typical services required for the vehicle number KE55KZB**

Considering another immediate risk vehicle number WR56NRJ, the total number of jobs is 10, the total labor hours is 23.7 and the average labor hours is 2.37. These results are shown in Figure 6. Hence it is required to allocate approximately 2.37 hours for this vehicle when scheduling appointment. The typical services required for this vehicle are shown in Figure 7. As obvious from the figure the typical services required for this vehicle are body exterior, electrical system and routine service repairs.

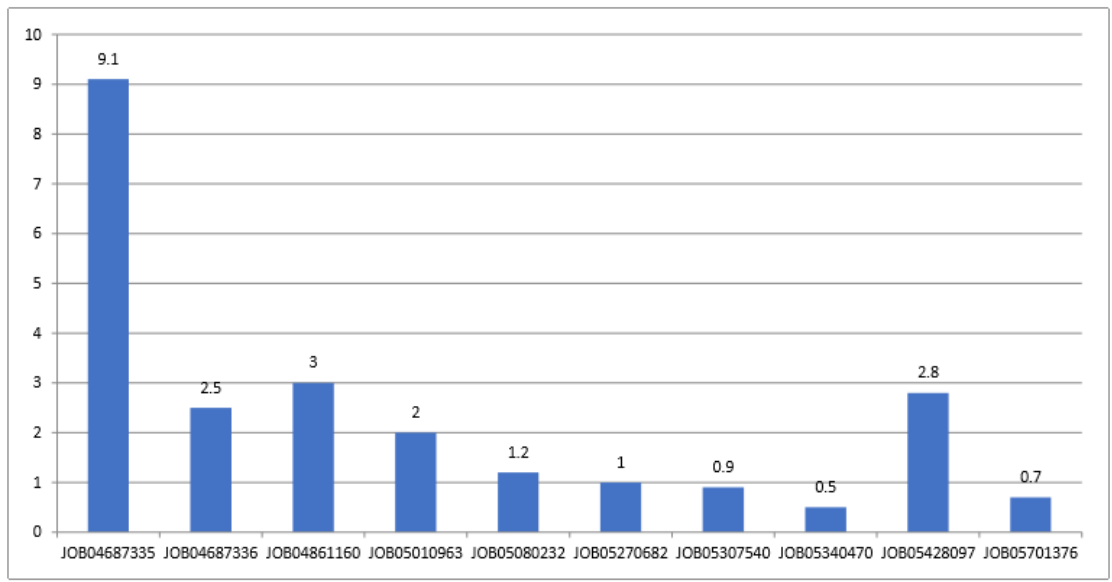

**Figure 6: Immediate risk vehicle number WR56NRJ (total number of jobs and labor hours)**

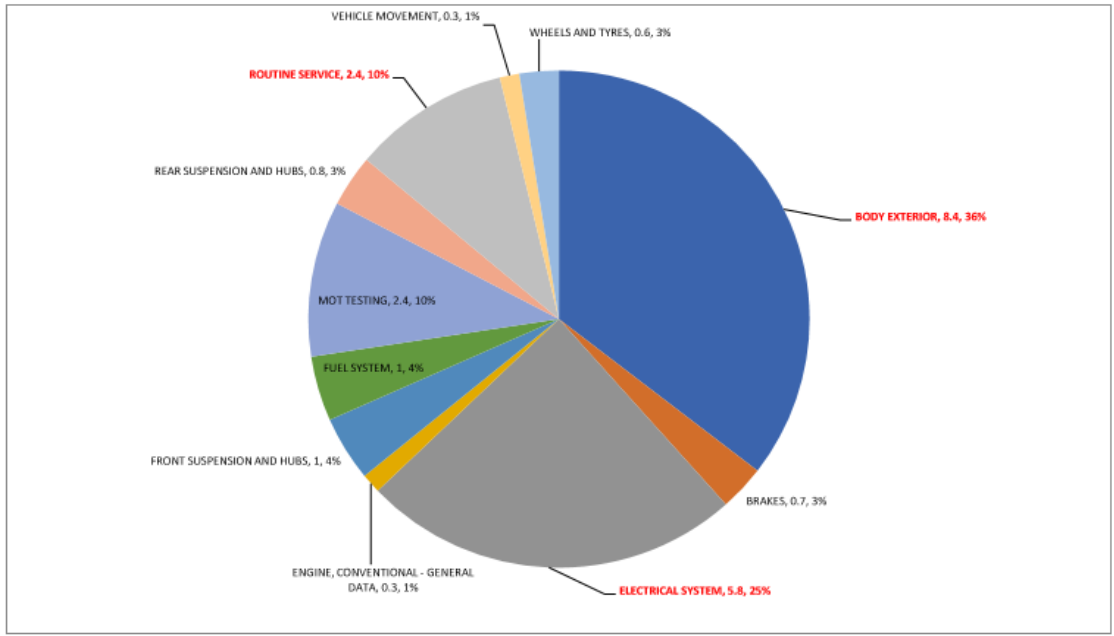

**Figure 7: The typical services required for the vehicle number WR56NRJ**

The receiver operating characteristic (ROC) curve or area under curve (AUC) for the predicted results from HMFSVM are shown in the Figure 8. The figure presents true positive versus false positive rates resulting from several cut-offs used in the prediction engine. It is better as the curve climbs faster. The ROC score obtained is 0.966 and the prediction accuracy is about 96%. The predictions being arrived at are in alignment with the longer-term objective of using real time sensor data for more accurate predictions.

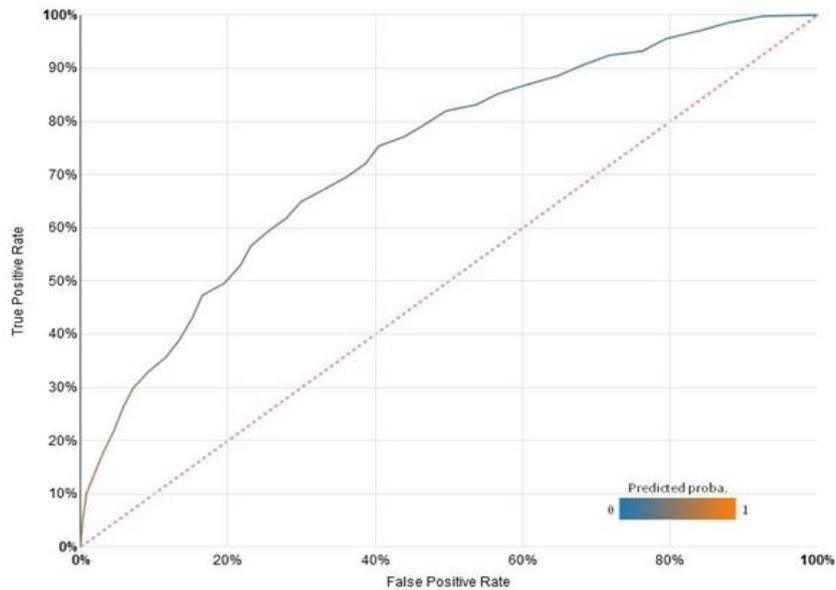

**Figure 8: The ROC curve for the predicted results**

HMFSVM has proved itself to be promising data classification approach. However, proper parameter selection is required to achieve good outcomes. In this direction cross validation finds the best combinations for parameters $C$ and $\gamma$ which are regularization and kernel parameters respectively. The best parameters are used for training and testing the whole dataset. Here hyperbolic tangent kernel is best suited based on which parameters $(C, \gamma)$ are chosen. This maps samples into higher dimensional space very easily. As such the best values of $(C, \gamma)$ for a given problem is not known beforehand. As a result of this, model selection or parameter search needs to be performed. This identifies best $(C, \gamma)$ values such the classifier accurately predicts the testing data. It is not always required to achieve high training accuracy. As a common strategy the dataset is separated into two parts comprising of one unknown component. The unknown component gives the prediction accuracy which gives the performance on classifying an independent dataset. This process is known as $v$ fold cross-validation where the training set is first divided into equal sized $v$ subsets. The classifier trained on the remaining $(v-1)$ subsets with the testing being performed on one subset. As a result of this each instance of the whole training set is predicted once. The cross-validation accuracy is correctly classified data percentage. The overfitting problem is prevented through this. Using cross-validation the grid search finds the values of $(C, \gamma)$. On trying exponentially growing $(C, \gamma)$ sequences good parameters are identified. The feasible values of $(C, \gamma)$ have the range $C = (2^{-5}, 2^{-3}, \ldots, 2^{18})$ and $\gamma = (2^{-18}, 2^{-17}, \ldots, 2^4)$. The motivation behind performing the grid search includes (a) it performs exhaustive parameter search using heuristics (b) it requires low computational times and (c) it can be easily parallelized because each $(C, \gamma)$ is independent. After populating the feasible region on the grid, a finer grid search finds the optimal values. A coarse grid is then used to find best $(C, \gamma)$ as $(2^8, 2^{-4})$ with cross validation rate 89 %. Next, finer grid search is conducted on the neighborhood of $(2^8, 2^{-4})$ such that better cross validation rate of 89.4 % at $(2^{8.45}, 2^{-4.45})$ is achieved. After that the best $(C, \gamma)$ is discovered on the whole training set to obtain the final classifier.

The Table 2 shows HMFSVM results as the number of support vectors on telecom-based company training and testing datasets with hyperbolic tangent kernel for different $C$ values.

| $C$ | Number of support vectors | Training rate (%) | Testing rate (%) |
|---|---|---|---|
| 8 | 14 (25.5 %) | 96.9 | 96.6 |
| 16 | 10 (19.4 %) | 96.9 | 96.6 |
| 64 | 10 (19.4 %) | 96.9 | 96.6 |
| 128 | 10 (19.4 %) | 96.9 | 96.6 |
| 256 | 8 (18.4 %) | 96.9 | 96.6 |
| 512 | 8 (18.4 %) | 96.9 | 96.6 |
| 1024 | 8 (18.4 %) | 96.9 | 96.6 |

Table 2: The results of HMFSVM with hyperbolic tangent kernel

HMFSVM reduces sensitiveness to class imbalance to a large extent which is present in SVMs **[26]**. This is achieved through algorithmic modifications to the HMFSVM learning algorithm. In lines with **[27]** a variant cost for the classes considered is assigned as misclassification penalty factor. A cost $\text{Cost}^+$ is used for the positive or minority class while and cost $\text{Cost}^-$ is used for the negative or majority class. This results in change to the fuzzy membership with a factor of $\text{Cost}^+\text{Cost}^-$. The ratio $\text{Cost}^+/\text{Cost}^-$ is the minority to majority class ratio **[28]** such that the penalty for misclassifying minority examples is higher. A grid search was performed to optimise the regularization and spread parameters. This has resulted in a model with the best sensitivity and without degrading specificity. HMFSVM is an optimization algorithm which reduces the support vectors considerably. This decreases the training times, simplifies model complexity with improved generalization.

Table 3 highlights performance of HMFSVM in terms of sensitivity, specificity, accuracy considering other algorithms. The support vectors in relation to training points $R_{sv/tr}$ is also presented for SVM based algorithms.

|  | logistic regression | random forest | SVM | FSVM | MFSVM | HMFSVM |
|---|---|---|---|---|---|---|
| sensitivity [%] | 65.86 | 69.86 | 79.86 | 85.89 | 89.86 | 95.66 |
| specificity [%] | 66.90 | 70.90 | 80.73 | 87.00 | 90.90 | 96.64 |
| accuracy [%] | 66.86 | 70.86 | 80.69 | 86.96 | 90.86 | 96.60 |
| $R_{sv/tr}$ |  |  | 0.909 | 0.954 | 0.969 | 0.986 |

**Table 3: The performance of HMFSVM with respect to other algorithms**

The Figure 9 highlights the performance of HMFSVM in terms of sensitivity, specificity and accuracy for different classification levels. The Figure 10 shows support vector ratios for different classification levels. Here the analysis is performed with respect to three ratios viz $R_{tr/max}$, $R_{sv/tr}$ and $R_{sv/max}$. The $R_{tr/max}$ denotes the training points in relation to maximum training points and $R_{sv/max}$ denotes the support vectors in relation to maximum training points.

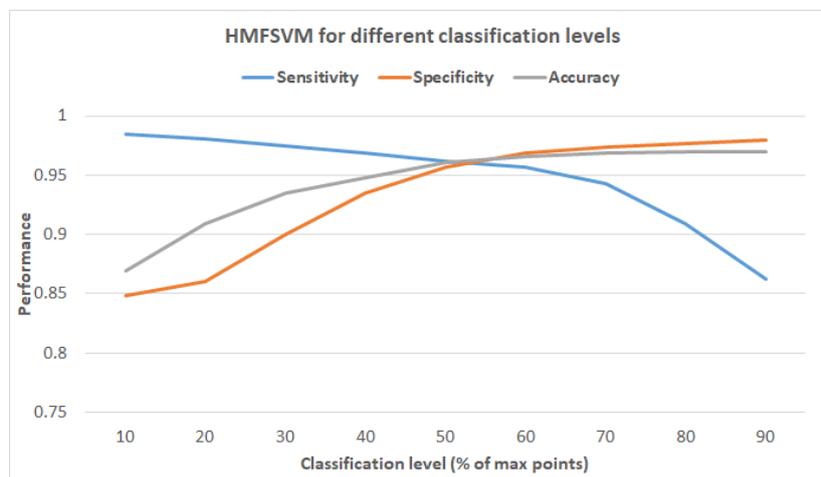

**Figure 9: The performance of HMFSVM for different classification levels**

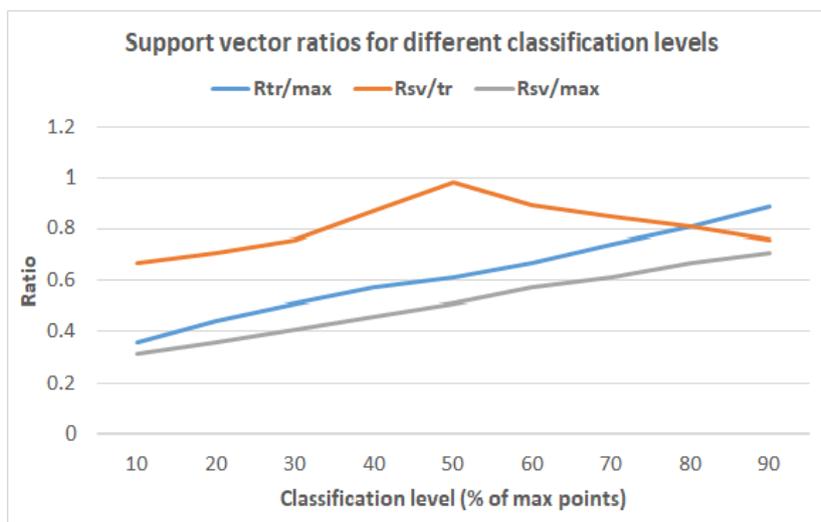

**Figure 10: The support vector ratios for different classification levels**

HMFSVM is also examined through McNemar test [29] which determines whether it significantly outperforms other techniques viz SVM, FSVM and MFSVM. This test is nonparametric in nature for related samples using chi-square distribution. McNemar test assesses significance of the difference between two dependent samples considering interesting variable value as dichotomy. It detects changes in responses considering intervention before and after designs [30]. The results of McNemar test are shown in Table 4. As evident from the Table 4 HMFSVM outperforms logistic regression and random forest at 5 % significance level and SVM, FSVM and MFSVM at 10 % significance level. The results highlight HMFSVM's effectiveness towards populating optimal feature subset and parameters through which the prediction results are improved significantly.

|  | random forest | SVM | FSVM | MFSVM | HMFSVM |
|---|---|---|---|---|---|
| logistic regression | 0.869 | 0.118 | 0.004** | 0.0005** | 0.0004** |
| random forest |  | 0.379 | 0.005** | 0.004** | 0.0005** |
| SVM |  |  | 0.086* | 0.069* | 0.064* |
| FSVM |  |  |  | 0.075* | 0.79* |
| MFSVM |  |  |  |  | 0.86* |

Table 4: McNemar test results on HMFSVM
(p-values for testing data: *Significant at 10 % level; **Significant at 5 % level)

We now highlight certain challenges faced during the development of the predictive maintenance prediction model [8]. A hypothesis was tried where the time differences between jobs was calculated. However the vehicle visits to the garages were irregular and sometimes within a week or even the next day of the earlier job. So this data could not be picked up for the hypothesis. Further individual task details for a job cannot be identified. For calculating vehicle age of the vehicle current date was considered. The vehicle age was calculated as current date – vehicle registration date. The vehicle failure binary variable was calculated using breakdown or callout. Only vehicles below 14 years of age was considered. The range of vehicles travelled considered was greater than 100 miles and less than 182000 miles.

## 10. Conclusion

Companies have sent field technicians on fixed schedules to perform routine diagnostic inspections and preventive maintenance on deployed vehicles. But this costly and labor intensive process. It does not ensure that failure won't happen between inspections. Predictive maintenance maximizes the vehicle lifespan and ensures optimized productivity. By harnessing the vehicle garage data several problems can be anticipated well ahead of failure. With the proposed hierarchical SVM-based solution for predictive maintenance organizations can rapidly deploy advanced analytics and gain the ability to make decisions in real time. Companies can conduct predictive maintenance and operations management with workflows triggered by the output of the analytics. With these real-time, actionable insights customers can make predictive maintenance a reality based on real-time analytics across their entire field deployment.